# Accounting for Uncertainty in Machine Learning Surrogates: A Gauss-Hermite Quadrature Approach to Reliability Analysis

**Amirreza Tootchi**

School of Mechanical Engineering

Purdue University

e-mail: atootchi@purdue.edu

**Xiaoping Du**

School of Mechanical Engineering

Purdue University

e-mail: duxp@purdue.edu

# Abstract

Machine learning surrogates are increasingly employed to replace expensive computational models for physics-based reliability analysis. However, their use introduces epistemic uncertainty from model approximation errors, which couples with aleatory uncertainty in model inputs–potentially compromising the accuracy of reliability predictions. This study proposes a Gauss-Hermite quadrature approach to decouple these nested uncertainties and enable more accurate reliability analysis. The method evaluates conditional failure probabilities under aleatory uncertainty using First- and Second-Order Reliability Methods and then integrates these probabilities across realizations of epistemic uncertainty. Three examples demonstrate that the proposed approach maintains computational efficiency while yielding more trustworthy predictions than traditional methods that ignore model uncertainty.

## 1. Introduction

Uncertainties significantly impact the reliability and safety of engineering systems, potentially leading to failures with severe consequences [1]. These uncertainties arise from inherent variability in material properties, operational conditions, and manufacturing processes (aleatory uncertainty) [2], as well as from incomplete knowledge and model approximations (epistemic uncertainty) [3,4]. For physics-based reliability analysis using computational models, accurately quantifying both types of uncertainty is essential for ensuring system safety across diverse applications, including aerospace structures, power systems, electromechanical systems, and civil infrastructure [5–9].

The First- and Second-Order Reliability Methods (FORM/SORM) have been widely adopted for efficient reliability analysis by approximating a computational model, called a limit-state function, through linearization or quadratic approximation at the Most Probable Point (MPP) [10–14]. However, for complex systems requiring expensive computational models like Finite Element Analysis (FEA) [15] or Computational Fluid Dynamics (CFD) [16], hundreds or thousands of model evaluations may be required, making reliability analysis impractical [17]. This computational challenge has driven the adoption of surrogate models, particularly those trained by

Machine Learning (ML) approaches such as Gaussian Process (GP) regression, which can approximate high-fidelity models at a fraction of the computational cost [18–21].

While surrogate models enable efficient reliability analysis, they introduce epistemic uncertainty from limited training data and model approximations [22,23]. This epistemic uncertainty couples with aleatory uncertainties from input variability, creating nested uncertainty structures that challenge traditional reliability methods [24–26]. Here, input variability refers to physical random quantities such as loading (axial force, bending moment, torque, pressure, thermal loads), material properties (Young's modulus, yield/ultimate strength, density), geometry & manufacturing tolerances (thickness, hole/shaft diameter, fillet radius, surface roughness, assembly clearances), and environment/operation (temperature, humidity, corrosion rate, friction coefficient, rotational speed). Recent studies have demonstrated that neglecting this coupling can lead to significant underestimation of probabilities of failure, particularly in high-reliability systems [27,28]. The coupled uncertainty appears as a "distribution of distributions", where the distribution parameters of the predicted response becomes uncertain, requiring specialized treatment beyond standard FORM/SORM approaches [29,30].

Recent approaches to handle coupled uncertainties include double-loop Monte Carlo methods and augmented-dimension techniques, but these suffer from computational inefficiency or loss of interpretability [31–33]. The need for efficient methods that properly decouple and propagate both uncertainty types through reliability analysis remains a critical gap in the literature.

This paper proposes a Gauss-Hermite quadrature approach that efficiently decouples aleatory and epistemic uncertainties in reliability analysis. The method extends classical FORM/SORM by: (1) computing the conditional probability of failure for fixed realizations of epistemic uncertainty, and (2) integrating these conditional probabilities using Gauss-Hermite quadrature weights over epistemic uncertainty. This avoids the computational burden of augmented-dimension approaches while maintaining mathematical rigor [22].

**The key contributions are twofold.** First, we develop a Gauss–Hermite–based decoupling strategy and corresponding computational algorithms that explicitly account for and decouple both aleatory uncertainty in model inputs and epistemic (model) uncertainty, leading to improved accuracy. The resulting capability enables more reliable decision making and engineering design by directly incorporating model uncertainty into the analysis. Second, we propose a framework to quantify the effect of both types of uncertainty on reliability analysis. This framework provides a basis for assessing the trustworthiness of reliability predictions, particularly when machine learning surrogates are employed. We then demonstrate the approach on three problems and compare it against methods that ignore model uncertainty.

The paper is organized as follows: Section 2 reviews relevant literature on reliability analysis and coupled uncertainty quantification. Section 3 presents the proposed methodology. Section 4 demonstrates the approach through engineering examples. Section 5 concludes with key insights and future directions.

## 2. Literature Review

Physics-based reliability analysis relies on computational models grounded in physics principles to predict reliability. Since computational models have errors, reliability analysis methods must address both aleatory uncertainty (inherent randomness) and epistemic uncertainty (knowledge limitations). This review explores three core areas: (1) traditional uncertainty quantification (UQ) and reliability analysis under aleatory uncertainty, (2) machine learning models and epistemic uncertainty, (3) UQ for nested, or coupled, aleatory–epistemic uncertainties. We then identify critical gaps in current methodologies that motivate the present research.

**2.1. UQ and reliability analysis for aleatory uncertainty**

Aleatory uncertainty—from inherent variability in loads, material properties, or environmental conditions—is a fundamental concern in reliability analysis. Central to this domain are β-index methods, including the Hasofer–Lind reliability index, which provides an invariant safety margin measure [34], and FORM, which locates the Most Probable Point (MPP) and applies a first-order linear approximation to estimate small probabilities of failure [35]. When computational models (limit-state functions) are highly nonlinear, SORM enhances accuracy using the curvature at the MPP [36]. Classic generalizations like the Rackwitz–Fiessler transformation accommodate non-normal inputs [37], and Breitung's asymptotic SORM further refines probability estimation using principal curvatures [36]. For systems with extremely low probabilities of failure, sampling-based methods can also be used, but computationally expensive. To remedy the problem, improved sampling methods have been developed. For example, subset simulation breaks a rare-event into a sequence of intermediate events and uses Markov Chain Monte Carlo (MCMC) for efficient sampling [38,39]; line sampling improves on FORM by averaging along an importance direction in the input space [40]; and adaptive importance sampling and sequential directional importance sampling techniques iteratively tune field direction or noise levels for rare-event contexts [30]. Hybrid approaches—including Kriging with subset simulation importance sampling, which combines adaptive Kriging surrogates with subset simulation—have also been proposed to reduce computational burden while maintaining accuracy [41]. Comparative studies benchmark FORM, SORM, subset simulation, response-surface methods, and line sampling across varying curvature and dimension regimes, demonstrating when each excels [42,43]. Much of the reliability literature addresses static, component-level problems with a single limit-state function. In this setting, surrogate models reduce computational cost but introduce epistemic uncertainty, so prior studies propagate both aleatory inputs and surrogate uncertainty while retaining classical FORM/SORM analyses. System-level reliability frameworks further extend FORM/SORM to account for component interactions and multi-mode dependencies [44]. Collectively, these methods form a structured suite for aleatory UQ—ranging from classical analytical methods to modern surrogate- and sampling-based strategies—each selected based on a trade-off between efficiency, accuracy, dimensionality, and practical applicability to safety-critical engineering reliability challenges [45].

**2.2. Machine learning models and model (epistemic) uncertainty**

The reliability methods discussed in Sec. 2.1 call computational models repeatedly, but these models, especially high-fidelity models (FEA, CFD, etc.), are expensive to evaluate. Surrogate models for computational models have therefore become prevalent for reducing computational costs. These surrogates (or metamodels) are data-driven approximations constructed using ML techniques, trained on a limited set of data points obtained from a finite number of evaluations of the underlying computational model. The surrogate models can then replace expensive simulations for tasks like design space exploration, reliability analysis, UQ, and optimization [1,46]. For example, running a computational model may require hours for each design iteration, whereas evaluating its surrogate counterpart takes only seconds [47,48].

However, a critical concern is the accuracy of these surrogate models. By construction, surrogates introduce model form error—the difference between surrogate predictions and the true model outputs—which can be estimated by epistemic uncertainty [23,46]. Unlike aleatory uncertainty stemming from variability, this uncertainty is due to lack of knowledge: limited training data, incomplete model structures, or simplifying assumptions. Notably, these prediction errors are typically unknown at new design points unless the original high-fidelity model is run, which defeats the purpose of using the surrogate. This challenge has been well recognized in the engineering design community: the surrogate's prediction error must be estimated by quantifying the model's epistemic uncertainty. In other words, along with a surrogate's point prediction, designers seek an uncertainty measure or confidence indicating the possible error in that prediction.

Substantial research in ML, reliability analysis, and reliability-based design has focused on UQ for surrogate models. In the ML domain, many algorithms inherently provide model uncertainty information [48,49]. For instance, a GP surrogate naturally yields a predicted distribution for the output, where the mean is the best estimate and the standard deviation indicates uncertainty. Advanced Bayesian neural network approaches and ensemble methods similarly can produce a mean and variance for predictions, offering a measure of confidence in the output [50].

Over the last decade, there has been a flurry of research on methods such as active learning (adaptive sampling) using model uncertainty to decide where new simulations should be run, multi-fidelity modeling that combines cheap low-fidelity models with expensive high-fidelity [10], ones to improve predictions efficiently [51], and enhanced reliability/robust design techniques that account for surrogate model uncertainty during optimization [52]. For instance, Li and Wang (2019) proposed techniques to quantify and reduce surrogate-induced uncertainty in reliability-based design optimization, ensuring that the optimized design remains feasible under model inaccuracies [1]. Mahadevan & Rebba (2006) quantified model-form and numerical solution errors and treated these errors as additional random quantities in reliability-based design optimization [53]. Recent surveys and tutorial papers provide overviews of these developments – e.g., Nemani et al. (2023) offer a comprehensive tutorial on UQ in machine learning models for reliability-based design, covering GP regression, Bayesian neural nets, ensemble methods, and metrics to assess predictive uncertainty [54]. These works underscore that accounting for surrogate model uncertainty is now recognized as essential for safety-critical or reliability-critical analysis. Despite substantial progress, effectively managing a surrogate's epistemic uncertainty alongside other uncertainties in a complex design process remains challenging.

**2.3. UQ for coupled aleatory–epistemic uncertainties**

In engineering systems, aleatory uncertainty often intersects with epistemic uncertainty, leading to nested or coupled uncertainty. An illustrative example is a GP surrogate for $g(\boldsymbol{X})$. Conditioned on a particular realization $\boldsymbol{X} = \boldsymbol{x}$, the GP predictive distribution of the output is $\mathcal{N}(\mu(\boldsymbol{X}), \sigma^2(\boldsymbol{X}))$, where $\mathcal{N}$ denotes a Normal (Gaussian) distribution with mean $\mu(\boldsymbol{X})$ and variance $\sigma^2(\boldsymbol{X})$. Here, $\mu(\boldsymbol{X})$ and $\sigma^2(\boldsymbol{X})$ are determined by the training data and kernel hyperparameters. Because $\boldsymbol{X}$ is random, $\mu(\boldsymbol{X})$ and $\sigma^2(\boldsymbol{X})$ are themselves random variables. Consequently, the system output exhibits a "distribution of distributions": for each realization of $\boldsymbol{X}$, the surrogate output is Gaussian with mean $\mu(\boldsymbol{X})$ and variance $\sigma^2(\boldsymbol{X})$, and these parameters vary with $\boldsymbol{X}$. Treating the surrogate prediction as deterministic ignores this coupling and can underestimate probabilities of failure.

Early approaches to this coupling were conservative: interval-based reliability methods [55] treated epistemic uncertainty with bounds or "worst-case" intervals alongside probabilistic inputs, ensuring safety under limited data. However, such methods often yield overly conservative designs. A more refined direction emerged with probabilistic treatment of epistemic uncertainty. Der Kiureghian [2,56] introduced a probabilistic framework for parameter uncertainty;

Sankararaman & Mahadevan [28,57] provided methods to decompose total uncertainty into aleatory and epistemic contributions, supporting more nuanced risk estimation.

Subsequent work introduced single-loop integration strategies that avoid nested Monte Carlo. Nannapaneni & Mahadevan [4] proposed a Bayesian reliability model that treats uncertain distribution parameters and model-form error as random variables embedded within a reliability index calculation—allowing aleatory and epistemic uncertainties to be handled in a unified simulation.

Du [22] presents a unified coupled MPP-based reliability method. It treats surrogate prediction error as an additional random variable and extends the MPP search to high-dimensional space, directly capturing the joint aleatory–epistemic probability of failure. Demonstrated on mechanical component design cases, the method shows that ignoring model error can lead to non-conservative (unsafe) designs, whereas accounting for it shifts the design toward greater safety margins.

## 2.4. Critical gaps

Although foundational approaches that integrate aleatory and epistemic uncertainties already exist [4,22,28,46,53], several practical challenges continue to hinder their widespread use in engineering applications. These challenges include the high computational cost of nested schemes when applied to large-scale problems; dimensional inflation and sensitivity to curvature when auxiliary variables are embedded in the most probable point (MPP) search; numerical instabilities arising when the surrogate's predictive variance is small; and, most critically, significant loss of accuracy when epistemic and aleatory uncertainties are indiscriminately treated as random variables in FORM and SORM. These issues limit the reliability of UQ in safety-critical engineering design.

This paper addresses the accurate issue by introducing a Gauss–Hermite quadrature approach that reduces computational burden while ensuring that surrogate model errors and input variability are jointly accounted for. Most importantly, our method directly targets the unresolved need to improve the accuracy of UQ for coupled uncertainties, providing a more reliable and tractable framework for safety-critical reliability analysis.

## 3. Methodology

This section presents a reliability procedure for coupled aleatory–epistemic uncertainty that preserves the standard FORM/SORM workflow and search dimension. We consider static, component-level problems with a single limit state. For each fixed realization of the surrogate's epistemic parameters, we evaluate the conditional probability of failure using FORM or SORM in the original input space—by this we mean the conditional cumulative distribution function (CDF) of the response evaluated at the failure threshold for that epistemic realization. We then integrate these conditional probabilities over the epistemic distribution using Gauss–Hermite quadrature to obtain both the average probability of failure ($p_f$) and its variability. By separating the MPP search from the epistemic integration, the approach avoids the dimensional inflation and curvature distortions of augmented-space methods and yields more accurate, robust estimates with only a small number of FORM/SORM evaluations.

### 3.1. Problem Formulation

Let the surrogate model be denoted as $Y = g(X)$. The goal of the UQ task is to compute the probability of failure for the response $Y$, given the probabilistic descriptions of both aleatory and epistemic uncertainties. Let $F_{Y|X}(y; p(X))$ be the conditional probability of failure distribution for $Y$ given $X$. $p(X)$ is a vector of distribution parameters that may vary with $X$. This distribution accounts for epistemic uncertainty, which arises from model errors. Let $f_Y(y)$ and $F_Y(y)$ be the probability density function (PDF) and probability of failure distribution for $Y$, respectively. Let the joint PDF of $X$ be $f_X(x)$.

Our goal is to find the overall distribution for $Y$ incorporates both epistemic uncertainty (via the conditional distribution $F_{Y|X}$) and aleatory uncertainty (via the distribution of $X$).

The joint PDF of $X$ and $Y$ is given by

$$f_{Y,X}(y, x) = f_{Y|X}(y|x) f_X(x) \tag{1}$$

And the PDF of $Y$ is

$$f_Y(y) = \int_{-\infty}^{+\infty} f_{Y,X}(y, x) dx = \int_{-\infty}^{+\infty} f_{Y|X}(y|x) f_X(x) dx \tag{2}$$

The probability of failure distribution for $Y$ is then computed as

$$F_Y(y) = \int_{-\infty}^{y} f_Y(y) = \int_{-\infty}^{y} \int_{-\infty}^{+\infty} f_{Y|X}(y'|x) f_X(x) dx dy' \tag{3}$$

Rewriting the order of integration:

$$F_Y(y) = \int_{-\infty}^{+\infty} \left[ \int_{-\infty}^{y} f_{Y|X}(y'|x) dy' \right] f_X(x) dx \tag{4}$$

Note that the inner integral defines the conditional CDF.

$$F_{Y|X}(y) = \int_{-\infty}^{y} f_{Y|X}(y'|x) dy' \tag{5}$$

Thus, the overall CDF of $Y$ is

$$F_Y(y) = \int_{-\infty}^{+\infty} F_{Y|X}(y) dx = \mathbb{E}_X[\Phi(F_{Y|X}(y))] \tag{6}$$

where $\mathbb{E}_X(\cdot)$ stands for expectation, and $\Phi(\cdot)$ is the standard normal cumulative distribution function.

Failure is the event that the response $Y$ falls at or below a chosen threshold $y$. The cumulative distribution function $F_Y(y) = \Pr(Y \leq y)$ therefore gives the probability of failure when evaluated at the threshold, $p_f = F_Y(y)$, and the reliability is $R = 1 - p_f$. Equations (3)–(4) show that $F_Y(y)$ can be written as an average of conditional distributions, $F_Y(y) = \int F_{Y|X}(y|x) f_X(x)\, dx = \mathbb{E}_X[F_{Y|X}(y|X)]$. In particular, the conditional probability of failure for a given input is $p_f = F_{Y|X}(y|x)$, and averaging $p_f(x)$ over the input distribution yields $p_f$.

In surrogate-based modeling, epistemic and aleatory uncertainties are inherently coupled because the surrogate model's prediction error (epistemic uncertainty) varies with the input location $X$ (which has aleatory uncertainty). When $X$ is near training points, the model is accurate and epistemic uncertainty is small; when $X$ is far from training data, epistemic uncertainty increases. This spatial dependence of model uncertainty on the random input creates a "probability of a probability" problem—meaning the probability of failure becomes a random quantity governed by another probability distribution. Such nested uncertainty structures complicate UQ and make traditional reliability methods inadequate. To address this, we adopt a decoupling strategy introduced in our previous work [22], which transforms the problem into a form where epistemic and aleatory uncertainties can be treated separately. This reformulation enables more tractable and interpretable reliability analysis without the complexities of second-order probability.

The decoupling approach defines a new random variable

$$Z = F_{Y|X}(Y; \mathbf{p}(X)) \tag{7}$$

By the probability integral transformation, $Z$ follows a uniform distribution on $[0,1]$, namely, $Z \sim \mathcal{U}[0,1]$. $Z$ is independent of $X$ and epistemic uncertainty.

The predicted response $Y$ can then be recovered by

$$Y = F_{Y|X}^{-1}(Z; \mathbf{p}(X)) \tag{8}$$

An important instance of this formulation arises when the surrogate is a GP model. In this case, the conditional distribution of $Y|X$ is Gaussian, given by

$$Y|X \sim \mathcal{N}(\mu_Y(X), \sigma_Y(X)) \tag{9}$$

Leading to

$$F_{Y|X}(y) = \Phi\left(\frac{y - \mu_Y(X)}{\sigma_Y(X)}\right) \tag{10}$$

and thus

$$F_Y(y) = \int_{-\infty}^{+\infty} \Phi\left(\frac{y - \mu_Y(x)}{\sigma_Y(x)}\right) f_X(x) dx = \mathbb{E}_X\left[\Phi\left(\frac{y - \mu_Y(X)}{\sigma_Y(X)}\right)\right] \tag{11}$$

To facilitate further analysis, we transform the standardized variable $Z$ into $U_Y$, which captures the combined effect of input randomness and surrogate predictive uncertainty.:

$$U_Y = \frac{y - \mu_Y(X)}{\sigma_Y(X)} \tag{12}$$

Accordingly, the response can be reformulated as

$$Y = \mu_Y(X) + U_Y \sigma_Y(X) \tag{13}$$

This reformulation enables the use of classical physics-based reliability analysis methods. In particular, FORM estimates the probability of failure by linearizing the limit-state surface at the

MPP. The reliability index, $\beta$, is defined as the minimum distance from the origin to the limit-state surface in the transformed standard normal space. The interpretation of $\beta$ as an invariant safety margin makes it a central measure in reliability analysis. Using FORM directly yields

$$F_Y(y) = \Phi(-\beta) \tag{14}$$

Its second-order counterpart, SORM, refines this by incorporating local curvature, yielding improved accuracy for nonlinear performance functions. Among notable SORM extensions, Breitung's formula offers an asymptotic correction using the principal curvatures of the limit-state surface, while Tvedt's correction accounts for higher-order derivatives to further reduce bias in cases of significant nonlinearity. These methods remain central in structural and mechanical reliability analysis; however, their effectiveness diminishes when probability boundaries are irregular or when surrogate predictive variance is small, conditions frequently encountered in surrogate-based UQ. Furthermore, when the standardized variable $Z$ is introduced to decouple input randomness from surrogate predictive uncertainty, additional approximation errors may arise. While this transformation simplifies the mathematical formulation, it also increases the dimensionality of the MPP search by adding one extra variable. This expansion can lead to reduced efficiency and degraded accuracy in locating the most probable point of failure, particularly in high-dimensional or strongly nonlinear problems, thereby amplifying the limitations of existing FORM/SORM-based approaches.

### 3.2. Gauss-Hermite Quadrature Reliability Method (GH-QRM)

Let $U_X \in \mathbb{R}^n$ denote the vector of standardized inputs in the standard normal space, obtained from the physical inputs via an isoprobabilistic transform $X = \boldsymbol{T}(U_X)$. Let $\mu_Y(\cdot)$ and $\sigma_Y(\cdot)$ be the surrogate's predictive mean and standard deviation evaluated at the physical inputs $T(U_X)$. We then introduce a standardized response variable $U_Y$ by writing the response as $y = \mu_Y[(T(U_X))] + U_Y \sigma_Y[T(U_X)]$.

The goal of the proposed Gauss-Hermite Quadrature Reliability Method (GH-QRM) is to enhance the accuracy of reliability analysis by addressing two primary drawbacks of the direct FORM and SORM approaches:

- The integration boundary in the standard normal space $U$ is irregular, as illustrated in Figure 1. This irregularity arises because the probability integration boundary is defined by

$$y = \mu_Y\big(T(U_{x_1}, U_{x_2})\big) + U_Y \sigma_Y\big(T(U_{x_1}, U_{x_2})\big) \tag{15}$$

which introduces peaks and discontinuities in the boundary surface.

- The standardization process in the probability transformation can result in extremely large values when the model's predictive variance is very small, leading to sharp peaks and numerical instability on the integration boundary.

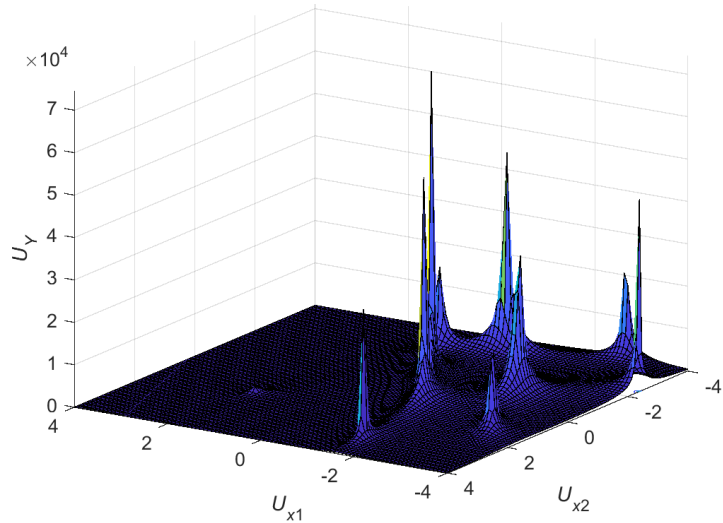

Figure 1. Probability integration boundary $y = \mu_Y\big(T(U_{x_1}, U_{x_2})\big) + U_Y \sigma_Y\big(T(U_{x_1}, U_{x_2})\big)$

Such limitations motivate two design objectives: (i) suppress sharp peaks along the probability-integration boundary to stabilize the MPP search; and (ii) avoid surrogate approximation in the auxiliary-variable space by expressing the boundary in terms of the standardized inputs $U_X$ and standardized response $U_Y$, which preserves predictive accuracy during reliability analysis.

The probability integration boundary is given by $y = \mu_Y[T(\boldsymbol{U}_X)] + U_Y \sigma_Y[T(\boldsymbol{U}_X)]$, which is rewritten as

$$U_Y = -\frac{\mu_Y[T(\boldsymbol{U}_X)]}{\sigma_Y[T(\boldsymbol{U}_X)]} \tag{16}$$

When the input $X$ or $\boldsymbol{U}_X$ is close to a training point, the GP model tends to be highly confident, resulting in a very small model error and thus a small $\sigma_Y$. As a result, the standardization in $U_Y = -\mu_Y[\boldsymbol{T}(U_x)/\sigma_Y[\boldsymbol{T}(U_x)]]$ leads to a very large magnitude for $U_Y$, even if $\mu_Y$ itself is moderate. This causes sharp peaks, as observed in Figure 1, and introduces numerical difficulties in the MPP search process in FORM/SORM.

To address this problem, we cap the value of $U_Y$ as follows: If $U_Y \geq 4$, we set $U_Y = 4$; if $U_Y \leq -4$, we set $U_Y = -4$. Since $\Phi(4) \approx 0.999968$ and $\Phi(-4) \approx 0.000032$, this cap bounds the probability of failure within the range $[0.000032, 0.999968]$. If a wider range is desired, a cap of 5 may be used. Figure 2 illustrates this capping on the probability integration boundary.

The capping does not affect the MPP search as long as the reliability index $\beta$ falls within the capped probability of failure range. This is because the MPP corresponds to the shortest distance $\beta$, and $\beta \leq U_Y$. Therefore, truncating values of $U_Y$ that exceed $\beta$ does not influence the result of the MPP search.

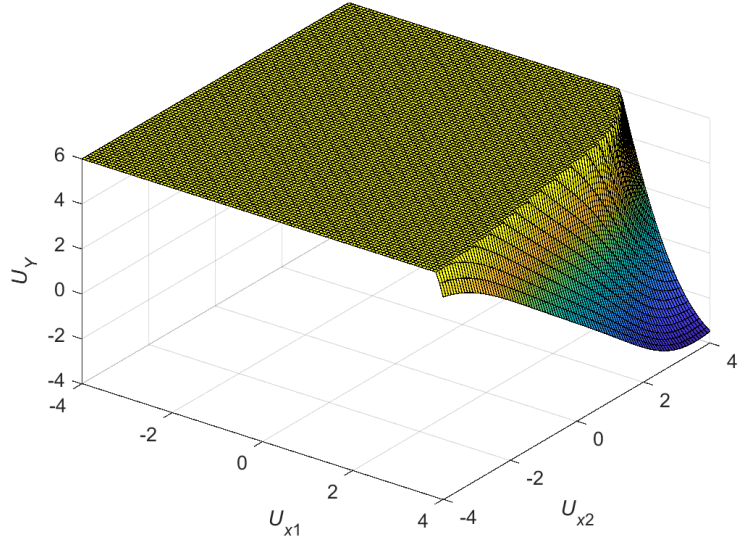

*Figure 2. Probability integration boundary after capping*

To address the approximation error introduced by the surrogate model during the standardization process involving $U_Y$ or $Z$, we propose a direct integration strategy that bypasses this step. Specifically, instead of including $U_Y$ or $Z$ in the FORM/SORM procedure, we evaluate the probability integrals directly over the original surrogate model output. This approach avoids approximating the Gaussian process surrogate with respect to $U_Y$, thereby preserving predictive accuracy in reliability analysis. Within this framework, the reliability estimates are obtained using enhanced variants of classical methods, namely FORM with Gauss Quadrature (FORM-GQ) and SORM with Gauss Quadrature (SORM-GQ), which combine the efficiency of reliability indices with the improved accuracy of Gauss–Hermite quadrature integration.

The joint PDF of $\boldsymbol{X}$ and $Z$ is

$$f_{X,Z}(\boldsymbol{x}, z) = f_X(\boldsymbol{x}) f_Z(z) = f_X(\boldsymbol{x}) \tag{17}$$

where $f_Z(z) = 1$ is the PDF of $Z$ since $Z \sim \mathcal{U}[0,1]$.

Using the recovered response in Eq. (8) with decoupled uncertainties, we have

$$F_Y(y) = \int_{-1}^{1} \int_{F_{Y|X}^{-1}(Z;\mathbf{p}(x))<0} f_{X,Z}(x,z)dx\,dz = \int_{-1}^{1} \left[\int_{F_{Y|X}^{-1}(Z;\mathbf{p}(x))<0} f_X(x)dx\right]dz \quad (18)$$

For a GP model, we replace $Z$ with $U_Y$. The joint PDF of $X$ and $U_Y$ becomes

$$f_{X,U_Y}(x,u_Y) = f_X(x)\phi(u_Y) \quad (19)$$

where $\phi(\cdot)$ is the probability of failure of a standard normal variable.

$$F_Y(y) = \int_{-\infty}^{+\infty} \int_{F_{Y|X}^{-1}(u_Y;\mathbf{p}(x))<y} f_{X,U_Y}(x,u_Y)dx\,du_Y = \int_{-\infty}^{+\infty} \left[\int_{F_{Y|X}^{-1}(u_Y;\mathbf{p}(x))<y} f_X(x)dx\right]\phi(u_Y)du_Y \quad (20)$$

Define

$$w(u_Y) = \int_{F_{Y|X}^{-1}(u_Y;\mathbf{p}(x))<y} f_X(x)dx \quad (21)$$

or

$$w(u_Y) = \int_{\mu_Y[T(U_X)]+U_Y\sigma_Y[T(U_X)]<y} f_X(x)dx \quad (22)$$

Then

$$F_Y(y) = \int_{-\infty}^{+\infty} w(u_Y)\phi(u_Y)du_Y \quad (23)$$

Evaluating $w(u_Y)$ involves computing the conditional probability of failure for $\mu_Y[T(U_X)] + U_Y\sigma_Y[T(U_X)]$ given $U_Y = u_Y$. This evaluation can then be performed using FORM/SORM without the approximation of the GP model with respect to $U_Y$. By integrating $w(u_Y)$ over $U_Y$, we obtain the probability of failure of $Y$. Therefore, the only source of error in using FORM/SORM arises from the approximation of the GP model with respect to only $\boldsymbol{X}$.

The integrand in Eq. (23) is the product of a nonlinear function $w(\cdot)$, obtained from FORM/SORM, and a standard normal density $\phi(\cdot)$. This type of integral is ideally suited for Gauss–Hermite quadrature, a numerical integration method specifically designed for integrals involving a standard normal weight function.

Gauss–Hermite quadrature provides high-accuracy approximation of integrals of the form

$$\int_{-\infty}^{+\infty} h(v)\phi(v)dv \approx \sum_{i=1}^{m} c_i h(v_i) \tag{24}$$

where $v_i$ are the quadrature points (roots of the Hermite polynomial), and $c_i$ are the corresponding weights. The method is especially efficient when the integrand is smooth and rapidly decaying, as is common in Gaussian-weighted integrals.

The higher accuracy of the proposed GH-QRM arises from the fact that Gauss–Hermite quadrature is specifically designed for integrals with Gaussian weights, which naturally align with the probability distributions used in reliability analysis. Unlike FORM/SORM approximations that rely on local linearization or curvature corrections near the MPP, the quadrature method directly integrates over the probability space without distortion from dimensional transformations or underestimation of model uncertainty. By exploiting well-chosen quadrature points and weights, GH-QRM achieves more accurate estimation of failure probabilities, especially in cases where surrogate predictive variance is small or probability boundaries are irregular. This ensures that both aleatory and epistemic uncertainties are captured with improved fidelity compared to traditional approaches.

### 3.2.1. Procedure of the Gauss-Hermite Quadrature Reliability Method (GH-QRM)

The procedure of the proposed method is summarized below:

**Step 1:** Determine the number of quadrature points $m$.

- Choose $m$ based on the desired accuracy. Common values range from 5 to 20.
- Higher $m$ gives better accuracy but increases computational cost or numbers of FORM/SORM calls.

**Step 2:** Loop over each quadrature point.

For each $i = 1,2,\ldots,m$,

- Obtain the quadrature point $v_i$ and corresponding weight $c_i$
- Call FORM using $u_Y = v_i$ to compute the conditional probability of failure $h(v_i) = w(u_Y)$ at $u_Y$.
- Multiply the result by the quadrature weight $c_i$.

**Step 3:** Compute final result.

Sum all weighted evaluations:

$$F_Y(y) = \sum_{i=1}^{m} c_i w(u_{Yi}) \tag{25}$$

### 3.3. Effect of model uncertainty

The probability of failure $F_Y(y)$ is deterministic if no model uncertainty exists. When model uncertainty presents (expressed by $Z$), conditional probability of failure $F_{Y|Z}(y)$ becomes random as $Z$ is random. For a GP model, model uncertainty is expressed by $U_Y$, and the conditional probability of failure is

$$F_{Y|U_Y}(y) = \int_{\mu_Y[T(U_X)]+U_Y\sigma_Y[T(U_X)]<y} f_X(x)dx = W(U_Y) \tag{26}$$

We predict the first two moments, or the mean and standard deviation of $W(U_Y)$. The mean of $W(U_Y)$ is give by

$$\mathbb{E}[W(U_Y)] = \int_{-\infty}^{+\infty} w(u_Y)\phi(u_Y)du_Y \tag{27}$$

which is given in Eq. (23) and can then be estimated using the GH-QRM in Eq. (24).

The variance of $W(U_Y)$ is given by

$$\mathbb{D}[W(U_Y)] = \int_{-\infty}^{+\infty} w^2(u_Y)\phi(u_Y)du_Y - \mathbb{E}^2[W(U_Y)] \tag{28}$$

Using the GH-QRM, we have

$$\mathbb{D}[W(U_Y)] \approx \sum_{i=1}^{m} c_i w^2(u_{Yi}) - \left[\sum_{i=1}^{m} c_i w(u_{Yi})\right]^2 \tag{29}$$

We can use the coefficient of variation (cov) to measure the effect of model uncertainty. The CV is calculated by $\mathbb{D}[W(U_Y)]$

$$cov = \frac{\sqrt{\mathbb{D}[W(U_Y)]}}{\mathbb{E}[W(U_Y)]} \tag{30}$$

The coefficient of variation ($cov$) in Eq. (30) quantifies the relative impact of model uncertainty on the probability of failure estimate. A larger $cov$ indicates greater variability in the predicted probability of failure due to model uncertainty, implying that the prediction is less reliable. In engineering design, a higher $cov$ suggests that model uncertainty plays a significant role in the analysis, and therefore more conservative decisions should be made. For instance, safety factors may be increased, or design margins may be widened to account for the higher level of uncertainty. Conversely, when $cov$ is small, the influence of model uncertainty is limited, and the decision-making process can be less conservative while still maintaining adequate reliability. Thus, Eq. (30) provides a useful metric to guide engineers in balancing safety, reliability, and efficiency under model uncertainty.

## 4. Examples

Three examples are presented. Example 1 is a one-dimensional problem. It is used for easy demonstration and virtualization. Example 2 involves an engineering problem with an analytical model and non-normal random input variables while Example 3 employs a block-box simulation model. The analytical models in the first two examples can be evaluated efficiently and repeatedly, making them well-suited for the comparison study.

In all examples, the probability of failure is defined as $p_f = P[Y < 0]$, where $Y$ is the limit state function. When $Y < 0$, the system is in the failure domain; when $Y > 0$, the system is safe. The boundary $Y = 0$ defines the limit state surface, separating safe and failure regions.

### 4.1. Example 1: A mathematical problem

The original model is analytical and is expressed as follows:

$$Y = 0.5(X + 4)^{0.5} - 0.58 \qquad (31)$$

where $X \sim N(0, 1^2)$.

**Case 1: Model (Epistemic) uncertainty is large**

A GP model is trained using three training points, as shown in Figure 3. The figure also shows the GP mean prediction, its 95% confidence band reflecting model uncertainty, and the true model. The objective is to determine the probability of failure for $Y$ at $X = 0$. From the perspective of reliability analysis, the task corresponds to predicting the probability of failure at the limit state $X = 0$, i.e., $p_f = P[Y < 0]$. As indicated in the figure, the model uncertainty is large at the limit state since the training points are far away from the limit state.

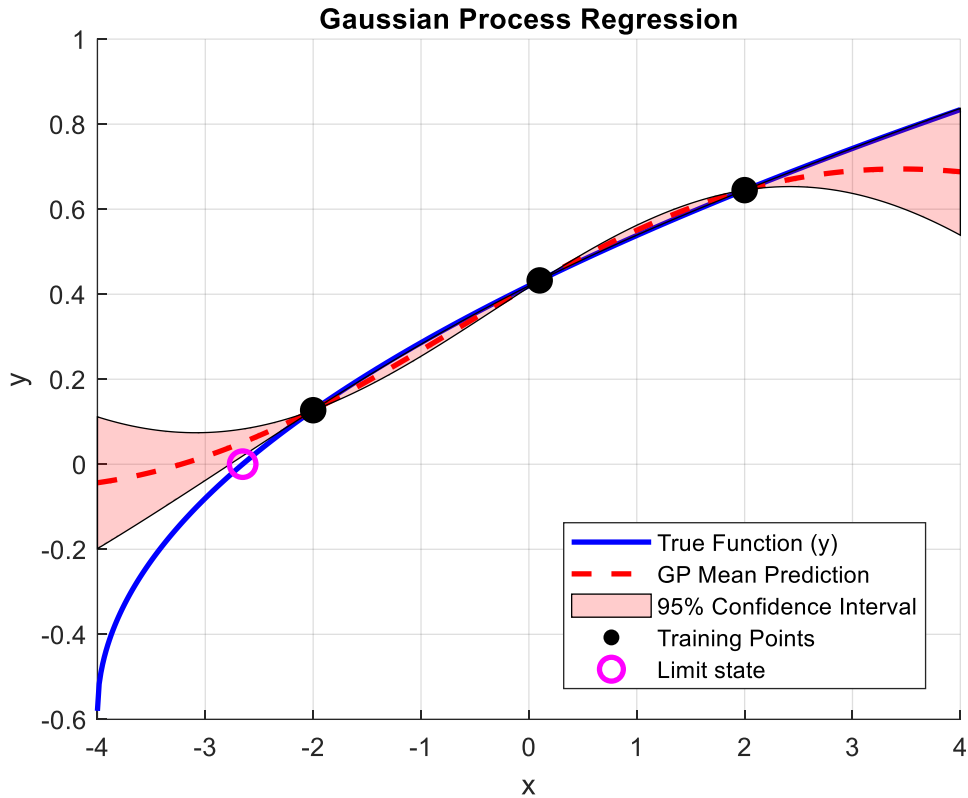

Figure 3. A GP Model Using Three Training Points

The probability of failure is

$$p_f = F_Y(0) = P[Y = \mu_Y(U_X) + U_Y \sigma_Y < 0] \qquad (32)$$

where $\mu_Y$ and $\sigma_Y$ are outputs of the GP model, and $U_X = X$.

The surface of $Y = \mu_Y + U_Y\sigma_Y$ and the contour $\mu_Y + U_Y\sigma_Y = 0$ are plotted in Figure 4, and the $\mu_Y + U_Y\sigma_Y = 0$ alone with clipping in the U-space is also plotted in Figure 5. The two figures illustrate that the $Y = \mu_Y + U_Y\sigma_Y$ is nonlinear with respect to $U_X$ and $U_Y$.

To demonstrate the benefits of considering model uncertainty, we perform UQ using direct FORM, the FORM-GQ, and MCS for the original and GP models. SORM does not work for this one-dimensional problem and is therefore not included. The reliability results for case 1 are summarized in Table 1. The effects of model uncertainty are also summarized in Table 2.

Table 1. Reliability results, for example 1 (Case 1) with significant model uncertainty

| Model | Method | $p_f$ | Error |
|---|---|---|---|
| Original model | Direct FORM | $3.9725 \times 10^{-3}$ | $-1.01\%$ |
| | MCS | $4.0130 \times 10^{-3}$ | — |
| GP model | Direct FORM | $1.0878 \times 10^{\wedge}(-3)$ | $28.27\%$ |
| | FORM-GQ | $8.3717 \times 10^{-4}$ | $-1.28\%$ |
| | MCS | $8.480 \times 10^{-4}$ | — |

Table 2. Effects of model uncertainty for example 1 (Case 1)

| Mean of the conditional $p_f$ (from FORM-GQ) | Standard deviation of the conditional $p_f$ (from FORM-GQ) | Coefficient of variation |
|---|---|---|
| $8.3717 \times 10^{-4}$ | $8.4285 \times 10^{-4}$ | $\approx 1$ |

**Observations:**

- The results of the CDF or probability of failure from the original and GP models are quite different due to large model uncertainty. The MCS solutions using the original and GP models are $4.0130 \times 10^{-3}$ and $8.480 \times 10^{-4}$, respectively. The latter is much smaller than the former, indicating the GP model is risky if it is used for reliability prediction.

- The direct FORM method is accurate when the original model, without any model uncertainty, is used, yielding a relative error of only $-1.01\%$. However, when significant model uncertainty is present, its accuracy degrades, and the relative error increases to 28.27%.

- The proposed method is much more accurate than the direct use of FORM. The relative errors with respect to the MCS solution are −0.28% and 28.27%, respectively.

- If we use the conditional $p_f$ which accounts for only aleatory uncertainty in model input, the effect of the model uncertainty is significant, indicated by a large $cov$ of approximately 1.

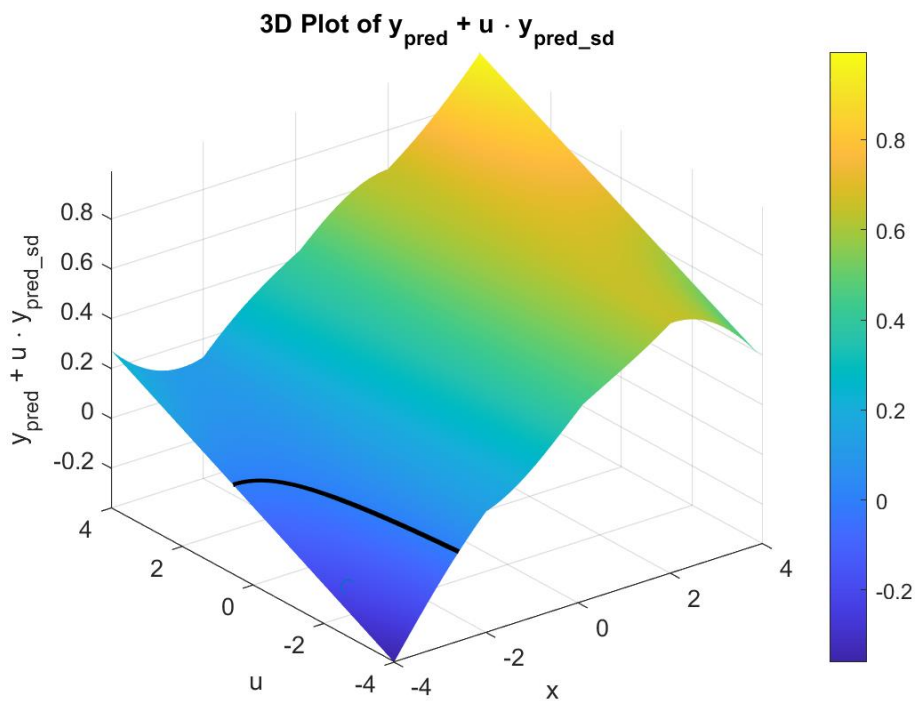

*Figure 4. Surface of $Y = \mu_Y + U_Y \sigma_Y$ with the contour $\mu_Y + U_Y \sigma_Y = 0$ over the $x - u$ domain*

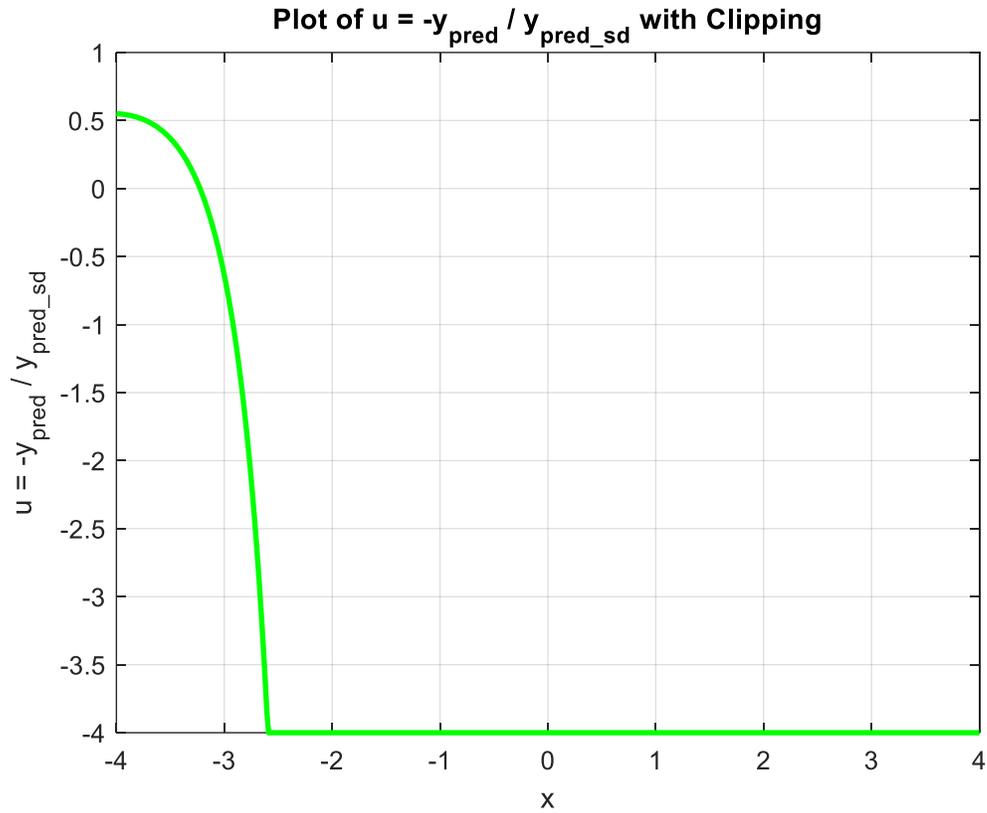

*Figure 5. Clipped curve of $\mu_Y + U_Y \sigma_Y = 0$ in the x domain*

**Case 2: Model uncertainty is small**

In this case, there are more training points and one of them is closer to the limit state; model uncertainty is less significant, as shown in Figure 6. The reliability results for case 2 are summarized in Table 3. Also, the result of the effect analysis of model uncertainty is summarized in Table 4.

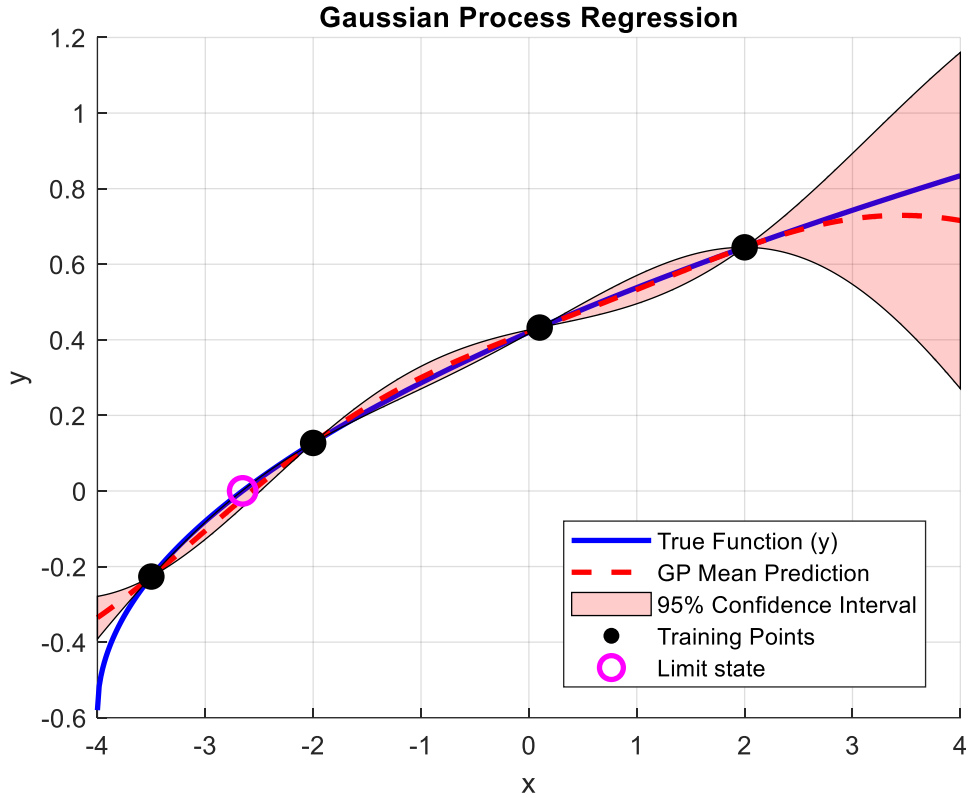

*Figure 6. GP prediction with small model uncertainty and a training point near the limit state.*

*Table 3. Reliability results, for example 1 (Case 2) when model uncertainty is insignificant*

| Model | Method | $p_f$ | Error |
|---|---|---|---|
| Original model | Direct FORM | $3.9725 \times 10^{-3}$ | $-1.01\%$ |
| | MCS | $4.0130 \times 10^{-3}$ | — |
| GP model | Direct FORM | $5.1550 \times 10^{-3}$ | $-0.22\%$ |
| | FORM-GQ | $5.1221 \times 10^{-4}$ | $-0.77\%$ |
| | MCS | $5.1430 \times 10^{-3}$ | — |

*Table 4. Effects of model uncertainty for example 1 (Case 2)*

| Mean of the conditional $p_f$ (from FORM-GQ) | Standard deviation of the conditional $p_f$ (from FORM-GQ) | Coefficient of variation |
|---|---|---|
| $5.1221 \times 10^{-4}$ | $6.3638 \times 10^{-4}$ | 0.1242 |

**Observations:**

- With insignificant model uncertainty, the prediction of $p_f$ from the GP model is close the that from the original model.

- When model uncertainty is insignificant, the relative error of the direct FORM is much smaller than that in Case 1. The relative error is only −0.22%.

- When model uncertainty is insignificant, the relative error of FORM-GQ is also small (−0.77%), though slightly larger than that of the direct FORM. However, the difference between the two methods is minor, and given that the comparison benchmark—MCS—also involves some randomness, it is difficult to definitively determine which method is more accurate.

- The uncertainty of the conditional $p_f$ is not significant, as its $cov$ is approximately 0.12—much smaller than that in Case 1, where the COV is around 1.0.

### 4.2. Example 2: Speed reducer shaft

A speed reducer shaft is subjected to a random force $P$ and a random torque $T$. The design model for yielding is defined by the difference between the yield strength and the maximum equivalent stress and is given by

$$Y = g(X) = S_y - \frac{16}{\pi d^3}\sqrt{4P^2 l^2 + 3T^2} \tag{33}$$

where $X = (S_y, d, l, P, T)$. The distributions and parameters of the basic random variables are described in Table 5.

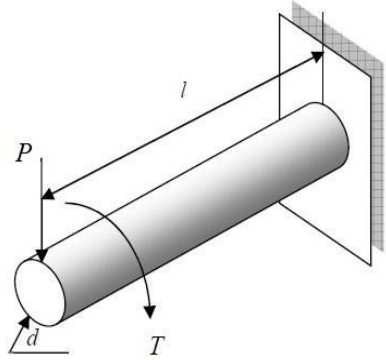

*Figure 7. Speed reducer shaft*

*Table 5. Distributions of the variables in speed reducer*

| Variables | Distribution | Mean | Standard Deviation |
|---|---|---|---|
| $S_y$ (*MPa*) | Lognormal | 250 | 30 |
| $d$ (*mm*) | Normal | 40 | 0.0001 |
| $l$ (*mm*) | Normal | 400 | 0.0001 |
| $P$ (*N*) | Normal | 1780 | 363 |
| $T$ (*N·m*) | Normal | 430 | 40 |

A GP model is built using 200 training points. The reliability results are summarized in Table 6. Also, the effects of model uncertainty are summarized in Table 7.

*Table 6. Reliability results, for example 2*

| Model | Method | $p_f$ | Error |
|---|---|---|---|
| Original model | Direct FORM | $8.8270 \times 10^{-3}$ | −13.46% |
| | SORM Breitung | $8.8270 \times 10^{-3}$ | −13.46% |
| | SORM Tvedt | $8.8270 \times 10^{-3}$ | −13.46% |
| | MCS | $1.020 \times 10^{-3}$ | — |
| GP model | Direct FORM | $1.3216 \times 10^{-3}$ | 8.32% |
| | FORM-GQ | $1.2968 \times 10^{-3}$ | 6.29% |
| | SORM-GQ Breitung | $1.3085 \times 10^{-3}$ | 7.25% |
| | SORM-GQ Tvedt | $1.2893 \times 10^{-3}$ | 5.68% |
| | MCS | $1.220 \times 10^{-3}$ | — |

Table 7. Effects of model uncertainty, for example 2

| Method | Mean of the conditional $p_f$ | Standard deviation of the conditional $p_f$ (from FORM-GQ) | Coefficient of variation |
|---|---|---|---|
| FORM-GQ | $1.3216 \times 10^{-3}$ | $3.8431 \times 10^{-4}$ | 0.30 |
| SORM-GQ (Breitung) | $1.3085 \times 10^{-3}$ | $3.6607 \times 10^{-4}$ | 0.28 |
| SORM-GQ (Tvedt) | $1.2893 \times 10^{-3}$ | $2.8394 \times 10^{-4}$ | 0.28 |

**Observations:**

- The probability of failure predicted by the GP surrogate model is slightly larger than that from the original model. According to Table 6, the Monte Carlo Simulation (MCS) result for the original model is $1.020 \times 10^{-3}$, while the GP model yields $1.220 \times 10^{-3}$. This indicates that the GP surrogate, despite being data-driven, introduces a mild overestimation in the probability of failure due to model form uncertainty.

- When applying the direct FORM method to the GP model, the relative error with respect to the MCS result increases to 8.32%, compared to the original model's FORM error of −13.46%, as shown in Table 6. This underscores that neglecting model uncertainty in reliability estimation leads to notable inaccuracies, even with a reasonably trained surrogate.

- In contrast, the proposed FORM-GQ method significantly improves the accuracy, reducing the relative error to 6.29%, while the SORM-GQ methods further improve performance. Specifically, the Breitung and Tvedt versions yield errors of 7.25% and 5.68%, respectively, suggesting that incorporating curvature improves prediction without substantially increasing computational cost.

- As detailed in Table 7, the coefficient of variation (cov) values for the conditional failure probabilities are 0.30 for FORM-GQ, and 0.28 for both SORM-GQ variants. These moderate values indicate that epistemic uncertainty is present and measurable, though not dominant. Compared to Example 1, where cov ≈ 1, the GP model in this example demonstrates better reliability due to more sufficient training coverage near the failure region.

- The small differences in conditional $p_f$ means and standard deviations across FORM-GQ and SORM-GQ methods (ranging from $1.2893 \times 10^{-3}$ to $1.3216 \times 10^{-3}$) support the robustness of the proposed approach. Moreover, all methods perform significantly better than direct FORM in terms of prediction accuracy.

- Overall, these results validate that the proposed Gauss-Hermite quadrature-based framework effectively accounts for nested aleatory and epistemic uncertainties, improving the credibility of reliability estimates derived from surrogate models, especially in moderately uncertain engineering problems like the speed reducer shaft.

### 4.3. Example 3: Deflection analysis of a bracket under load

In Example 3, a bracket component subjected to mechanical loading, where a GP surrogate replaces expensive FEA simulations. The test case is adapted from the *Deflection Analysis of a Bracket* example provided by MathWorks [58], ensuring consistency with established finite element benchmarks.

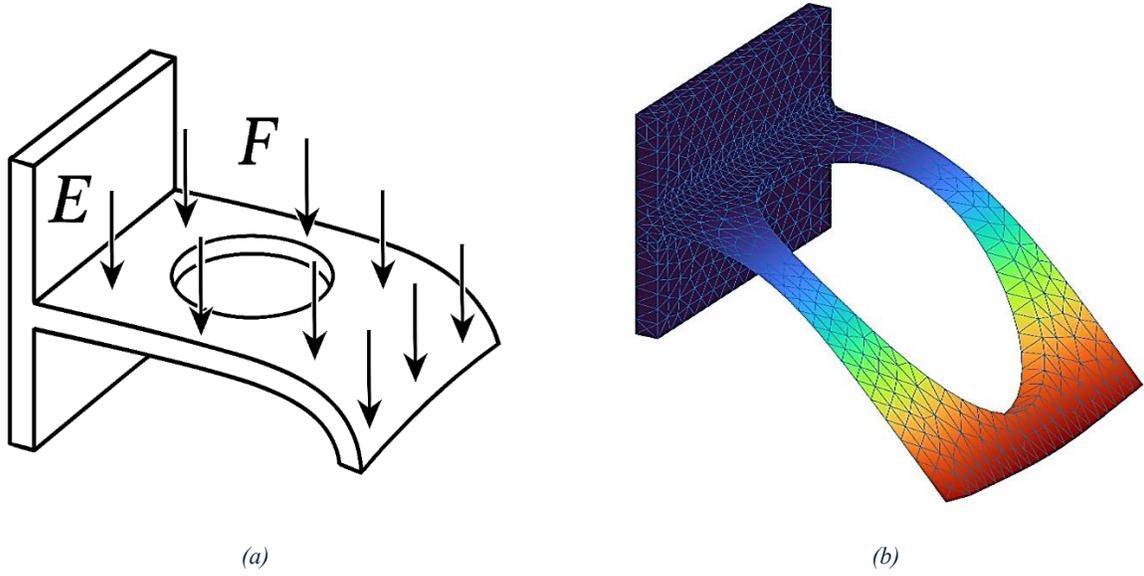

*(a)* *(b)*

*Figure 8. (a) Bracket schematic under load, (b) Finite-element deflection results*

The failure criterion is defined based on the vertical deflection at the free end of the bracket. The critical deflection threshold $\delta_{cr}$ is chosen based on engineering judgment and corresponds to the maximum allowable displacement before performance degradation or structural failure occurs. In this study, the critical deflection threshold is defined as $\delta_{cr} = \frac{L}{85}$, where $L$ is the bracket's effective length, representing a practical limit for allowable displacement based on structural design considerations. The limit-state function is given by

$$Y = \delta_{cr} - \delta_z(X) \tag{34}$$

where $\delta_z(X)$ is the deflection and is evaluated by FEA. $X = [E, F]$. Failure occurs when $Y < 0$.

Two random variables—Young's modulus $E$ and applied distributed force $F$—introduce aleatory uncertainty. Their distributions are provided in Table 8. These variables serve as the inputs to the GP surrogate model, while the output is the predicted bracket displacement, accompanied by a predictive variance representing epistemic uncertainty. This output is then integrated via GH-

QRM to explicitly account for epistemic uncertainty. Reliability is subsequently evaluated using FORM-GQ and SORM-GQ, with MCS.

The GP model is trained on N=10 data points sampled over the joint uncertainty space of $E$ and $F$, with corresponding outputs representing the true deflections obtained from high-fidelity FEA. The surrogate captures both the nonlinearity of the response and the influence of geometry, including the circular cutout in the bracket. This GP model is then embedded into FORM-GQ and SORM-GQ methods for probability of failure estimation.

Table 8. Distributions of the variables in bracket under load

| Variables | Distribution | Mean | Standard Deviation |
|---|---|---|---|
| $E\ (GPa)$ | Lognormal | 260 | 26 |
| $F\ (N/m)$ | Normal | -504000 | 16500 |

Unlike Example 1 and Example 2, the original FEA model is not used for the comparison study due to its high computational cost. Instead, the analysis compares two GP-based approaches for modeling uncertainty. Both approaches employ the same GP surrogate models to replace computationally intensive finite element analyses. Z-based methods refers to our previous auxiliary-variable framework for separating aleatory and epistemic uncertainties [22], while the GQ-based methods implements the newly proposed GH-QRM formulation. This ensures a consistent comparison of epistemic uncertainty modeling strategies within the same GP surrogate setting. The reliability results are summarized in Table 9.

Table 9. Reliability results, for example 3

| Model | Method | $p_f$ | Error (%) |
|---|---|---|---|
| GP model-Z | Direct-FORM | 0.001809 | 17.09 |
| | Direct-SORM Breitung | 0.001779 | 15.16 |
| | Direct-SORM Tvedt | 0.001785 | 15.54 |
| | MCS-Z | 0.001545 | – |
| GP model-GQ | FORM-GQ | 0.001564 | 1.47 |

|  | SORM-GQ Breitung | 0.001552 | 0.71 |
|---|---|---|---|
|  | SORM-GQ Tvedt | 0.001579 | 2.43 |

*Table 10. Effects of model uncertainty, for example 3*

| Method | Mean of the conditional $p_f$ | Standard deviation of the conditional $p_f$ (from FORM-GQ) | Coefficient of variation |
|---|---|---|---|
| FORM-GQ | $1.5636 \times 10^{-3}$ | $6.0490 \times 10^{-3}$ | 3.8686 |
| SORM-GQ (Breitung) | $1.5520 \times 10^{-3}$ | $6.0632 \times 10^{-3}$ | 3.9067 |
| SORM-GQ (Tvedt) | $1.5785 \times 10^{-3}$ | $6.2069 \times 10^{-3}$ | 3.9321 |

**Observations**

- The reliability results in Table 9 demonstrate that the GQ-based method consistently outperforms the Z-based method across all evaluation techniques, including FORM and both SORM variants. The superior accuracy of the GQ approach is evident in its consistently lower relative errors when benchmarked against the MCS-GQ result, which is considered the reference ($p_f=1.541 \times 10^{-3}$).

- In FORM analysis, the Direct-FORM yields a probability of failure of $p_f=1.809 \times 10^{-3}$, resulting in a relative error of 17.09%, whereas the FORM-GQ reduces this error dramatically to 1.47% with $p_f=1.564 \times 10^{-3}$. This significant improvement reflects the more accurate handling of epistemic uncertainty in the GQ approach, which integrates over both aleatory and epistemic domains via quadrature.

- For SORM-Breitung, the Direct-SORM Breitung result is $p_f=1.779 \times 10^{-3}$ with a 15.16% error, while the SORM-GQ Breitung achieves $p_f=1.552 \times 10^{-3}$, lowering the error to just 0.71%. Similarly, in SORM-Tvedt, the SORM-GQ Tvedt achieves an error of 2.43%, versus

15.54% for Direct-SORM Tvedt. These consistent improvements confirm the GQ method's robustness in capturing tail behavior under uncertainty.

- As detailed in Table 10, the cov of the conditional failure probabilities is 3.8686 for FORM-GQ, 3.9067 for SORM-GQ (Breitung), and 3.9321 for SORM-GQ (Tvedt). These are large cov values ($> 1$), indicating substantial relative dispersion in $p_f$ across Gauss–Hermite nodes and a strong influence of epistemic/model uncertainty in Example 3. In comparison, Example 1 (cov ≈ 1) and Example 2 (cov ≈ 0.3) show much lower relative variability. Note that a large cov reflects variability relative to the mean $p_f$; it does not contradict the small integrated $p_f$ reported for this example.
- While the Z-based methods use an auxiliary variable to treat epistemic uncertainty indirectly, it introduces approximation errors due to linearization and moment-matching, which accumulate especially in nonlinear systems. In contrast, the GQ-based methods apply numerical integration directly over the joint probability space, preserving nonlinear interactions and avoiding approximation bias.
- The consistent advantage of GH-QRM across all reliability techniques further justifies its use as the preferred method in UQ, especially when computational cost is manageable via surrogate modeling.

## 5. Conclusion

This study presents a mathematically rigorous and computationally tractable framework for reliability analysis in surrogate-based modeling under coupled aleatory and epistemic uncertainties. By extending classical FORM and SORM techniques through two complementary strategies—auxiliary-variable embedding and Gauss–Hermite quadrature—the proposed methodology enables efficient and accurate estimation of failure probabilities without compromising fidelity or interpretability. The quadrature-based decoupling, in particular, preserves the probabilistic structure of model uncertainty while avoiding the dimensional complexity typically associated with nested formulations. Benchmark applications demonstrate that neglecting model-form uncertainty can significantly underestimate risk, whereas the proposed framework delivers robust reliability estimates even in data-scarce regimes. Importantly, the approach is generalizable, making it broadly applicable to surrogate models built using Gaussian processes, neural networks, or other machine learning architectures. Future research will extend the proposed method to engineering design problems characterized by multiple responses or multiple surrogate models, with the central challenge being the identification of the joint distribution of all outputs under uncertainty. Incorporating this framework into design will enable system-level reliability analysis in complex, high-dimensional domains, where both aleatory and epistemic uncertainties play critical roles. This direction opens new theoretical and computational avenues for advancing robust and reliable engineering design.